\documentclass{article} 
\usepackage{iclr2027_conference,times}

\usepackage{amsmath,amsfonts,bm}

\def\eqref#1{equation~\ref{#1}}

\def\1{\bm{1}}

\DeclareMathAlphabet{\mathsfit}{\encodingdefault}{\sfdefault}{m}{sl}
\SetMathAlphabet{\mathsfit}{bold}{\encodingdefault}{\sfdefault}{bx}{n}

\usepackage{hyperref}
\usepackage{graphicx}
\usepackage{url}
\usepackage{booktabs}
\usepackage{multirow}
\usepackage{colortbl}
\usepackage{bm}
\usepackage{tabularx}
\usepackage{array}
\usepackage{xcolor}
\usepackage{soul}
\usepackage{siunitx}
\usepackage{enumitem}
\usepackage{wrapfig}
\usepackage{hhline}
\usepackage{makecell}
\usepackage{algorithmic}
\usepackage{wrapfig}
\usepackage{algorithm}
\usepackage{multirow}
\usepackage{siunitx} 
\usepackage{caption}
\usepackage{wrapfig}

\definecolor{mygray}{gray}{0.9}
\definecolor{myred}{RGB}{192, 0, 0}
\definecolor{mygreen}{RGB}{0, 176, 80}

\newcommand{\x}{\bm{x}}
\newcommand{\xt}{\bm{x}_t}
\newcommand{\xT}{\bm{x}_T}
\newcommand{\xzero}{\bm{x}_0}
\newcommand{\y}{\bm{y}}
\newcommand{\h}{\bm{h}}

\newcommand{\grad}{\nabla_{x_t}\log p_t}
\newcommand{\ywave}{\widetilde{\bm{y}}}
\newcommand{\ewave}{\widetilde{\bm{\epsilon}}}
\newcommand{\s}{\text{s}_{\theta}}
\newcommand{\hgt}{\textcolor{myred}{\h_{\xzero=\y}}}
\newcommand{\happ}{\textcolor{mygreen}{\h_{\xzero=\ywave}}}

\usepackage{color}
\definecolor{citecolor}{RGB}{66,168,235}
\definecolor{linkcolor}{RGB}{255,0,0}
\hypersetup{colorlinks=true,citecolor=citecolor,linkcolor=linkcolor}

\title{Adaptive Weighted $h$-Transform Sampling for Coarse-Guided Visual Generation}

\author{Yanghao Wang$^\star${\quad}Ziqi Jiang$^\star${\quad}Zhen Wang{\quad}and Long Chen$^\dagger$
\\
The Hong Kong University of Science and Technology
}

\newcommand{\ie}{\textit{i.e.}}
\newcommand{\eg}{\textit{e.g.}}
\newcommand{\cf}{\textit{c.f.}}

\iclrfinalcopy 
\begin{document}

\maketitle

\begin{abstract}
Coarse-guided visual generation, which synthesizes fine visual samples from degraded or low-fidelity coarse references, is essential for various real-world applications. While training-based approaches are effective, they are inherently limited by high training costs and restricted generalization due to paired data collection. Accordingly, recent training-free works propose to leverage pretrained diffusion models and incorporate guidance during the sampling process. 
However, these training-free methods either require knowing the forward (fine-to-coarse) transformation operator, \eg, bicubic downsampling, or are difficult to balance between guidance and synthetic quality. To address these challenges, we propose a novel guided method by using the $h$-transform, a tool that can constrain stochastic processes (\eg, sampling process) under desired conditions.
Specifically, we modify the transition probability at each sampling timestep by adding to the original differential equation with a drift function $h$, which approximately steers the generation toward the ideal fine sample.
To address unavoidable approximation errors, we introduce an adaptive weight scheduler that combines a noise-level-aware initialization with a correction based on cross-timestep consistency, balancing guidance adherence and synthesis quality. Extensive experiments across diverse image and video generation tasks demonstrate its effectiveness and generalization. \href{https://github.com/HKUST-LongGroup/Coarse-guided-Gen}{Code}.
\end{abstract}

\section{Introduction}
\label{sec:1}
Recent advances in diffusion~\citep{ho2020denoising,song2020score} and flow matching~\citep{liu2022flow,lipman2022flow} models have demonstrated remarkable visual generation capability. Benefiting from large-scale pretraining, high-quality samples can be generated by existing unconditional and text-to-visual (T2V) models~\citep{rombach2022high,flux2024,wan2025wan} in the case of no guidance or only with text guidance. However, some real-world applications (\eg, deblurring and super-resolution) always require a coarse visual sample to guide synthesis, which these pretrained unconditional and T2V models cannot handle directly. Thus, recent works~\citep{song2020score,choi2021ilvr,chung2022diffusion,kawar2022denoising,chung2022improving, kim2025flowdps,meng2021sdedit,singer2025time} aim to solve such a \textbf{coarse-guided visual generation} task: given a \emph{coarse visual sample} (\eg, a blurred image, or a warped video) as the guidance, the model needs to generate the \emph{corresponding refined sample} (\eg, a clear image, or an undistorted video).

\begin{figure*}[t]
    \centering
    \includegraphics[width=0.9\linewidth]{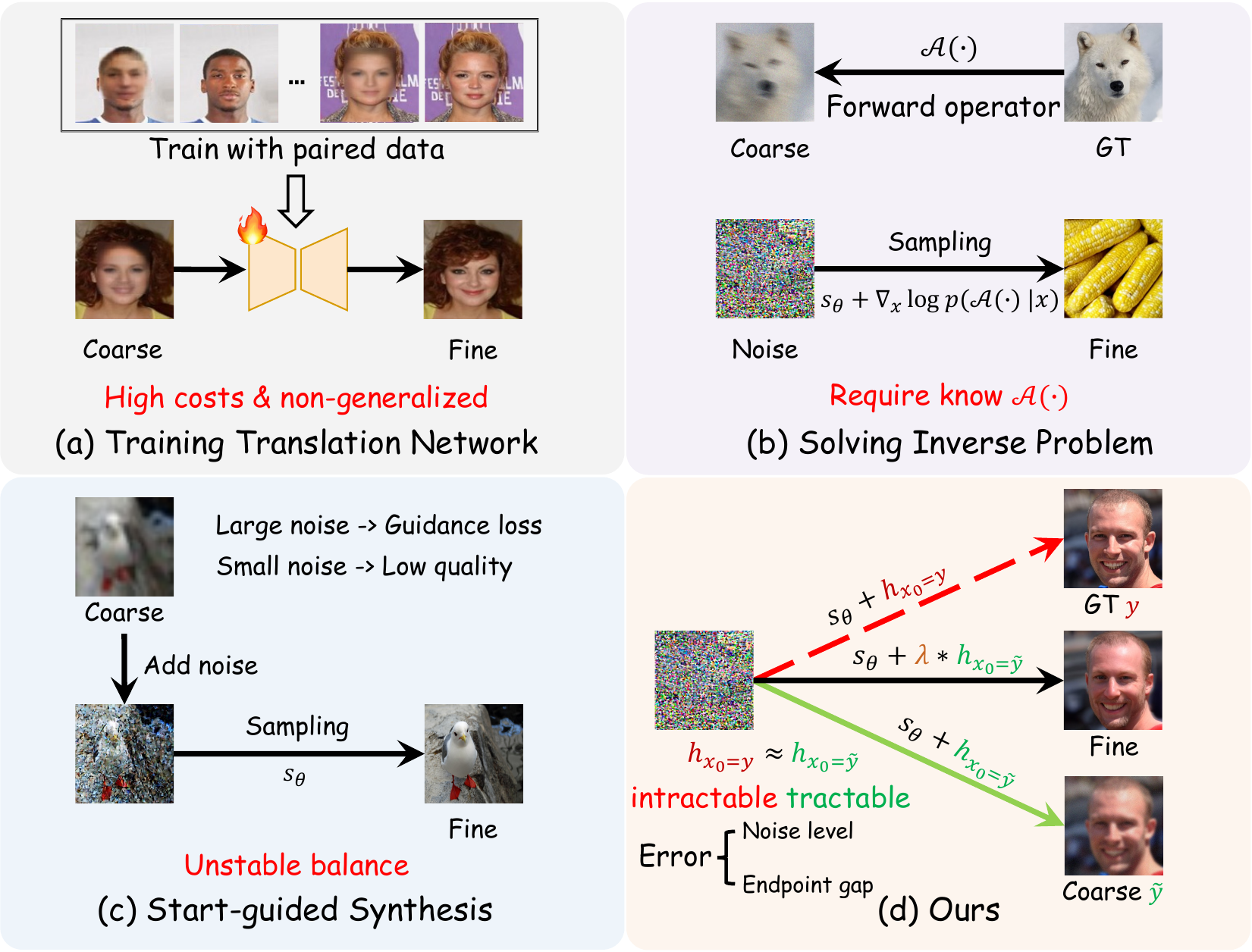}
    \caption[]{\textbf{Existing and our solutions.} (a) Training translation networks based on paired data, which is costly and non-generalizable to different types of coarse samples. (b) Solving inverse problems based on a known forward operator, which is not robust. (c) Adding noise to the coarse sample and denoising it, which is difficult to balance guidance and quality. (d) Our method leverages the $h$-transform to achieve training-free, operator-free, and stable coarse-guided generation.}
    \vspace{-1em}
    \label{fig:2}
\end{figure*}

A straightforward approach is to train translation networks. Shown in Fig.~\ref{fig:2}(a), previous studies~\citep{zhou2023denoising, de2021diffusion, zhang2023adding} try to train from scratch or fine-tune a translation model with paired coarse-fine data. For different types of coarse samples (\eg, low-resolution images, blurred images, and warped videos), additional/existing models must be trained/fine-tuned with the corresponding paired data. Thus, these training-based approaches suffer from inherent \textit{high costs} (\ie, training and data collection) and \textit{limited generalization}.

Subsequent work resorts to training-free solutions that only modify the sampling process to achieve guided generation. 
There are two mainstream training-free strategies: 
1) \textbf{Solving Inverse Problem}~\citep{song2020score,choi2021ilvr,chung2022diffusion,kawar2022denoising,chung2022improving, kim2025flowdps}:
As shown in Fig.~\ref{fig:2}(b), instead of sampling from the marginal distribution, they approximate the posterior probability term and sample directly from the conditional distribution. However, the approximation used in these methods usually relies on a known forward fine-to-coarse operator, \ie, the transformation from fine samples to coarse samples must be known, such as bicubic downsampling or a Gaussian mask operator. Thus, this prior (known operator) requirement reduces the robustness of these methods, as the operator may be unknown in many scenarios.
2) \textbf{Start-Guided Synthesis}~\citep{meng2021sdedit,singer2025time}: As shown in Fig.~\ref{fig:2}(c), they first get the sampling start by adding noise to the coarse sample. Then they can sample from it with a pretrained diffusion model to get the refined sample. However, the guidance is entirely dependent on setting the noisy coarse sample as the starting point. Adding a large amount of noise results in loss of guidance signals, whereas adding a small amount of noise yields only limited quality improvement. 
In that case, this start-dependent guided manner leads to an \textit{unstable balance} between guidance faithfulness and generation quality.

Based on the above discussions, we observe that current solutions are constrained by training, the requirement for known operators, and an unstable balance. 
To this end, inspired by Doob's $h$-transform~\citep{doob1984classical,rogers2000diffusions}, a tool that can constrain the sampling process with the given guidance, we propose a novel coarse-guided generation method, \textbf{Adaptive Weighted $h$-Transform Sampling}. Intuitively, our method incorporates guidance signals during sampling by modifying the transition probability at each time step and finally generates a guided refined result. This probability modification can be interpreted as introducing additional traction toward the \emph{ideal underlying refined result} on top of the original sampling route.

Specifically, we achieve traction by adding a new drift adjustment $\hgt$ ($\x$ is the time-index updating latent and $\y$ is the ideal result) to the original predicted score $\s$ (\cf, Figure~\ref{fig:2}(d)).
By sampling with a new score ($\s+\hgt$), the result can be guaranteed to be $\y$. However, this $\hgt$ is not tractable because it depends on knowing the ground truth, \ie, the ideal result $\y$. Since we can not directly obtain $\hgt$, we found that we can leverage another tractable $\happ$ to approximate it, where $\ywave$ is the given coarse sample. Subsequently, we analyze the approximation error and find that it depends on two factors:
1) It is negatively correlated with the \textbf{noise level} of $\x$. To mitigate the error influence, we use a noise-level-aware schedule as initialization to adjust the weight of $\happ$ across time steps. That is, as the approximation error increases, we decrease the weight of $\happ$ during the sampling process.
2) It is proportional to the \textbf{endpoint gap} (the distance between ideal $\y$ and coarse $\ywave$). Thus, we build a cross-time consistency criterion as the endpoint gap proxy to further adjust the noise-level-aware initialization into a better adaptive weight for each timestep.
In this way, the generated sample not only adheres to the coarse guidance but also improves in quality.



Our contributions are threefold: 1) We propose \emph{Adaptive Weighted $h$-Transform Sampling}, a training-free method for coarse-guided visual generation. It derives a tractable coarse-endpoint approximation to the ideal $h$-transform and requires no known forward degradation operator. 2) To mitigate the influence of approximation error, we factor the error into the noise level and unknown endpoint gap. Based on these two guidelines, we design an adaptive weight schedule that contains a noise-level-aware initialization complemented by a consistency correction. 3) Extensive experiments across different image and video generation tasks show our effectiveness and generalization.

\section{Background}
\label{sec:2}

\subsection{Diffusion Models}
\label{sec:2.1}
Diffusion models capture the target data distribution $p_{0}$ by learning a transport process from a prior distribution $p_{T}$ (\eg, a Gaussian) to $p_{0}$. To achieve that, a forward diffusion process from $p_{0}$ to $p_{T}$ is first determined and can be described with such a stochastic differential equation (SDE):
\begin{equation}
\label{eq:1}
    \mathrm{d}\x = \textbf{f}(\xt,t)\mathrm{d}t + g(t)\mathrm{d}\textbf{w},
\end{equation}
where $t\in[0,T]$ is the time-index, $\xzero \sim p_{0}$, $\textbf{f}$ is the drift function, $g$ is the diffusion coefficient, and $\textbf{w}$ is a Brown motion. 
Meanwhile, the marginal distribution of $\xt$ at time $t$ corresponding to this SDE is $p_t(\xt)$. Then, we have a reversed SDE to describe the reversed process (from $p_{T}$ to $p_{0}$):
\begin{equation}
\label{eq:2}
    \mathrm{d}\x = [\textbf{f}(\xt,t)-g^2(t)\grad(\xt)]\mathrm{d}t + g(t)\mathrm{d}\overline{\textbf{w}},
\end{equation}
where $\overline{\textbf{w}}$ is the reversed Brown motion.
According to the Fokker–Planck equation~\citep{maoutsa2020interacting} (F-P Equation), we can convert this reversed SDE into an ordinary differential equation (ODE) that has the same marginal distribution $p_t(\xt)$:
\begin{equation}
\label{eq:3}
    \mathrm{d}\x = [\textbf{f}(\xt,t)-\frac{1}{2}g^2(t)\grad(\xt)]\mathrm{d}t. 
\end{equation}
We can sample synthetic samples by solving the SDE in Eq.~(\ref{eq:2}) or the ODE in Eq.~(\ref{eq:3}). However, $\grad(\xt)$ (also known as the score) is not known. Thus, a parameterized network $s_{\theta}(\cdot)$ with parameters $\theta$ can be trained to predict it. The optimization target is:
\begin{equation}
\label{eq:4}
    \mathop{\mathrm{min}}_{\theta} \mathbb{E}_{\xt,\xzero,t} \left[||s_\theta(\xt,t)-\grad(\xt|\xzero)||_2^2 \right],
\end{equation}
where $\xzero\sim p_0$, $t \in [0,T]$ and $\xt\sim p_t(\xt|\xzero)$. $\grad(\xt|\xzero)$ is \textbf{tractable} since the conditional distribution is defined by the forward SDE in Eq.~(\ref{eq:1}), and it has closed-form solutions. After training, $s_{\theta}$ can be used as a score predictor to replace $\grad(\xt)$ for solving Eq.~(\ref{eq:2}) or Eq.~(\ref{eq:3}).

\subsection{Doob's $h$-Transform}
\label{sec:2.2}

For a given SDE like Eq.~(\ref{eq:1}), the corresponding process can transport $\xzero \sim p_0$ to Gaussian noise. However, in some situations, such as image translation, we may need to establish a mapping relation between $\xzero$ and a fixed point $\y$. Thus, Doob’s $h$-transform~\citep{doob1984classical,rogers2000diffusions} proposes to modify the probability transition of the original SDE by adding a $g^2(t)\h_{\xT=\y}$ term to the original drift term $\textbf{f}$:
\begin{equation}
\label{eq:5}
    \mathrm{d}\x = [\textbf{f}(\xt,t)+g^2(t)\h_{\xT=\y}]\mathrm{d}t + g(t)\mathrm{d}\textbf{w},
\end{equation}
where $\h_{\xT=\y}$ is called the $h$ function. The new transition guarantees that the SDE ends at the fixed point $\y$,
where $\h_{\xT=\y}$ is designed as $\grad(\xT=\y|\xt)$. It improves the transition probability of $p_t(\xT=\y|\xt)$ and guarantees $\xT = \y$ by adjusting the drift coefficient across timesteps.

\section{Adaptive Weighted $h$-Transform Sampling}
\label{sec:3}

\begin{figure*}[t]
    \centering
    \includegraphics[width=1\linewidth]{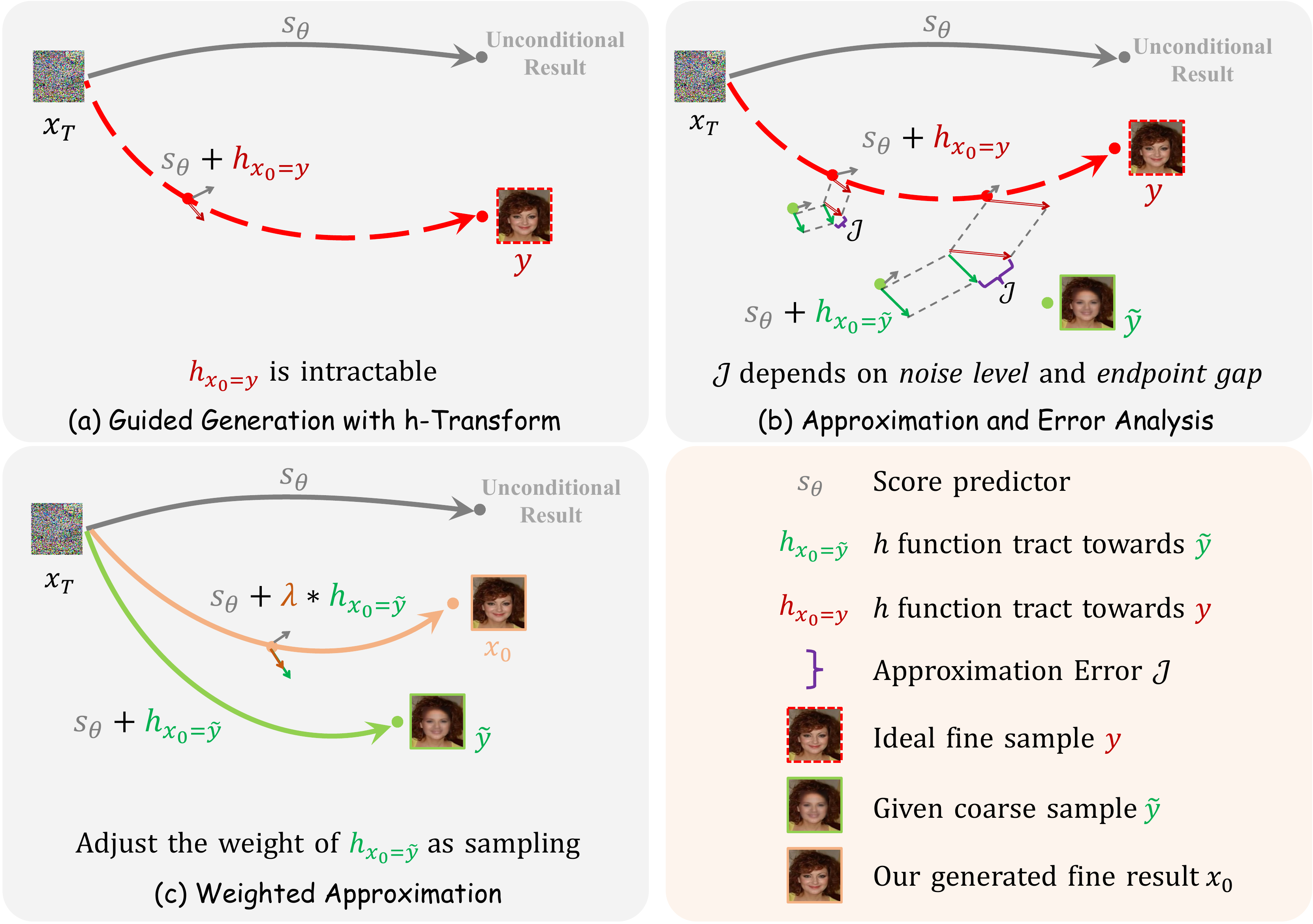}
    \vspace{-2em}
    \caption[]{\textbf{Overview of Weighted $h$-Transform Sampling}. (a) If we have $\hgt$, the generation result will be the ideal sample. (b) We leverage $\happ$ to approximate the intractable $\hgt$ and derive that the error depends on the noise level and endpoint gap. (c) To mitigate the error influence, we adjust the approximation weight and finally generate a high-quality refined sample.}
    \vspace{-1em}
    \label{fig:3}
\end{figure*}

\noindent\textbf{Problem Formulation.} Assuming that an underlying joint distribution $(\ywave, \y) \sim q$ describes the probability measurement between a coarse visual sample $\ywave$ and its corresponding fine sample $\y$. For the coarse-guided visual generation task, given a coarse sample as the guidance $\ywave$, our objective is to leverage a pretrained diffusion model $\s$ to generate a corresponding fine sample.
\subsection{Guided Generation with $h$-Transform}
If the ideal fine sample $\y$ were known, Doob's $h$-transform (Section~\ref{sec:2.2}) would condition the reverse process on $\xzero=\y$. Define $\hgt=\nabla_{\xt}\log p_t(\xzero=\y\mid\xt)$. The resulting reverse SDE is
\begin{equation}
\label{eq:6}
    \mathrm{d}\x=\left[\textbf{f}(\xt,t)-g^2(t)\bigl(\grad(\xt)+\hgt\bigr)\right]\mathrm{d}t+g(t)\mathrm{d}\overline{\textbf{w}},
\end{equation}
where time runs from $T$ to $0$. With the exact conditional score, this process is a bridge to $\y$, as illustrated in Figure~\ref{fig:3}(a). Its probability-flow ODE is
\begin{equation}
\label{eq:7}
    \mathrm{d}\x=\left[\textbf{f}(\xt,t)-\frac{1}{2}g^2(t)\bigl(\grad(\xt)+\hgt\bigr)\right]\mathrm{d}t.
\end{equation}
The SDE and ODE share conditional marginals when initialized from the same conditional distribution. However, the required drift adjustment $\hgt$ depends on the unknown target $\y$, making this ideal construction unavailable for coarse-guided generation.

\section{Approximating the Intractable \texorpdfstring{$\hgt$}{hgt}}
As $\hgt = \grad(\xzero=\y|\xt)$ is intractable, we manage to convert it into a tractable and appropriate approximation. To this end, we propose to approximate it with the following:
\begin{equation}
\begin{aligned}
\label{eq:8}
    \hgt=\grad(\xzero=\y|\xt) & \approx \grad(\xzero=\ywave|\xt).
\end{aligned}
\end{equation}
We name $\grad(\xzero=\ywave|\xt)$ as $\happ$, which is another $h$ function (\cf, Figure~\ref{fig:3}(b)). Then, we use Bayes' rule to change $\happ$ into:
\begin{equation}
\begin{aligned}
\label{eq:9}
\happ &=
    \grad(\xzero=\ywave|\xt)  = \grad(\xt|\xzero=\ywave) - \grad(\xt),
\end{aligned}
\end{equation}
where the second term $\grad(\xt)$ can be replaced by the given trained score predictor $\s$.
Besides, we already known the conditional distribution $p_t(\xt|\xzero)$:
\begin{equation}
\begin{aligned}
\label{eq:10}
    p_t(\xt|\xzero) = \mathcal{N}(\xt; \alpha_t\xzero, {\sigma_t}^2\textbf{I}),
\end{aligned}
\end{equation}
where $\alpha_t$ and ${\sigma_t}$ (noise level) are determined by $\textbf{f}(\xt,t)$, $g^2(t)$ of Eq.~(\ref{eq:1}). When $\xzero=\ywave$, which is also given as the coarse sample, $p_t(\xt|\xzero=\ywave)$ is closed-form as $\mathcal{N}(\xt; \alpha_t\ywave, {\sigma_t}^2\textbf{I})$ and $\nabla_x\log p_t(\xt|\xzero=\ywave)$ is tractable:
\begin{equation}
\begin{aligned}
\label{eq:11}
    \nabla_x\log p_t(\xt|\xzero=\ywave) &= 
    \nabla_{\xt}\left(-\frac{1}{2\sigma_t^2} (\xt - \alpha_t\ywave)^\textbf{T}(\xt - \alpha_t\ywave)\right) \\
    &= \frac{1}{\sigma_t^2} (\alpha_t\ywave-\xt).
\end{aligned}
\end{equation}
After that, we convert the intractable $\hgt$ into the following tractable approximation:
\begin{equation}
\begin{aligned}
\label{eq:12}
    \hgt \approx \happ = \frac{1}{\sigma_t^2} (\alpha_t\ywave-\xt) - \grad(\xt).
\end{aligned}
\end{equation}

\subsection{Approximation Error Analysis}
\label{sec:3.3}

Our proposed approximation in Eq.~(\ref{eq:8}) inevitably introduces approximation errors. We can calculate the Euclidean distance between $\hgt$ and $\happ$ as the approximation error $\mathcal{J}$: 
\begin{equation}
\begin{aligned}
\label{eq:13}
    \mathcal{J} &= ||\nabla_x\log p_t(\xzero=\y|\xt) - \nabla_x\log p_t(\xzero=\ywave|\xt)||_2 \\
    &=||\nabla_x\log p_t(\xt|\xzero=\ywave) - \nabla_x\log p_t(\xt|\xzero=\y)||_2 = ||\frac{\alpha_t}{\sigma_t^2}(\ywave - \y)||_2.
\end{aligned}
\end{equation}
Taking the variance-preserving (VP) diffusion model as the example ($\alpha_t^2+\sigma_t^2=1$), we can get:
\begin{equation}
\begin{aligned}
\label{eq:14}
    \mathcal{J} &= ||\frac{\sqrt{(1-\sigma_t^2)}}{\sigma_t^2}(\ywave - \y)||_2.
\end{aligned}
\end{equation}
We can see that the first factor of approximation error $\mathcal{J}$ is negatively correlated with the \textit{noise level} $\sigma_t$. Meanwhile, when $\sigma_t \rightarrow 0$, $\mathcal{J}\rightarrow \infty$ and when $\sigma_t \rightarrow 1$, $\mathcal{J}\rightarrow 0$. Besides, the error is also proportional to the second factor $\|\ywave-\y\|_2$, \ie, the \textit{endpoint gap}.

\subsection{Adaptive Weighted Approximation for Error Control}
\label{sec:3.4}
We weight the approximate adjustment by a scalar $\lambda_t\in[0,1]$, giving the guided score $\s(\xt,t)+\lambda_t\happ$. In the ODE notation of Eq.~(\ref{eq:7}), this yields the controlled sampling dynamics:
\begin{equation}
\label{eq:15}
    \mathrm{d}\x=\left[\textbf{f}(\xt,t)-\frac{1}{2}g^2(t)
    \left(\s(\xt,t)+\lambda_t\left(\frac{\alpha_t\ywave-\xt}{\sigma_t^2}-\s(\xt,t)\right)\right)\right]\mathrm{d}t.
\end{equation}
\noindent\textbf{Noise-level-aware initialization.}
The noise-dependent amplification in Eq.~(\ref{eq:13}) motivates a smooth weight function $\lambda_{\sigma,t}\in[0,1]$ of the noise level $\sigma_t$. We typically parameterize this schedule as a power function of $\sigma_t$. As the noise level decreases and the approximation error grows, the weight gradually decreases, with $\lambda_{\sigma,t}\to1$ as $\sigma_t\to1$ and $\lambda_{\sigma,t}\to0$ as $\sigma_t\to0$. This retains coarse guidance when the approximation is reliable and reduces its influence as the error is amplified.

\noindent\textbf{Consistency for endpoint gap.}
To account for the unknown endpoint gap, we use prediction consistency as a model-based proxy and define the consistency residual:
\begin{equation}
\label{eq:16}
\bm{r}_t(\lambda)
=\widehat{\bm{x}}_{0|t^-}^{(\lambda)}
-\widehat{\bm{x}}_{0|t}^{(\lambda)}.
\end{equation}
where $t^-$ denotes the next sampling time. The current guided clean prediction is
\begin{equation}
  \widehat{\bm{x}}_{0|t}^{(\lambda)}
  =(1-\lambda)\widehat{\bm{x}}_{0|t}^{(0)}+\lambda\ywave,
  \qquad
  \widehat{\bm{x}}_{0|t}^{(0)}
  =\frac{\xt+\sigma_t^2\s(\xt,t)}{\alpha_t}.
\end{equation}
Meanwhile, $\widehat{\bm{x}}_{0|t^-}^{(\lambda)}$ is the model's clean prediction after taking one sampling step with weight $\lambda$, without additional coarse guidance. A smaller residual norm indicates stronger agreement between the guided endpoint and the model's next-step prediction.
On top of that, we evaluate two weights: the initialized $\lambda_{\sigma}$ and a nearby $\lambda^{'}_{\sigma}\in[0,1]$, perturbed toward the interval interior. Let $\bm{r}_{\sigma},\bm{r}^{'}_{\sigma}$ denote their residual vectors in
Eq.~(\ref{eq:16}). A local linear approximation of the residual gives the damped least-squares update~\citep{marquardt1963algorithm}:
\begin{equation}
\label{eq:17}
\begin{aligned}
  \Delta\lambda
  &=-\frac{(\lambda^{'}_{\sigma}-\lambda_{\sigma})
  \langle\bm{r}^{'}_{\sigma}-\bm{r}_{\sigma},\bm{r}_{\sigma}\rangle}
  {(1+\kappa)\|\bm{r}^{'}_{\sigma}-\bm{r}_{\sigma}\|_2^2},\quad
  \lambda_t=\!\left(\lambda_{\sigma}+\Delta\lambda\right).
\end{aligned}
\end{equation}
Here, $\kappa\geq0$ controls damping.
The resulting sampler combines noise-aware initialization with consistency-based weight refinement, as summarized in Algorithm~\ref{Alg:1}.


\begin{algorithm}[!t]
\caption{Adaptive Weighted $h$-Transform Sampling}
\label{Alg:1}
\begin{algorithmic}[1]
\REQUIRE Coarse sample $\ywave$, score model $\s$, weight schedule
$\lambda_{\sigma}$, damping $\kappa$.
\STATE Initialize $\bm{x}_T\sim p_T$.
\FOR{each sampling time $t$}
\STATE Set $\lambda_t=\lambda_{\sigma}$ and choose a nearby
$\lambda'_\sigma\in[0,1]$.
\STATE $\bm{r}_\sigma=
\widehat{\bm{x}}_{0|t^-}^{(\lambda_\sigma)}
-\widehat{\bm{x}}_{0|t}^{(\lambda_\sigma)}$,
$\quad\bm{r}'_\sigma=
\widehat{\bm{x}}_{0|t^-}^{(\lambda'_\sigma)}
-\widehat{\bm{x}}_{0|t}^{(\lambda'_\sigma)}$.
\STATE $\displaystyle
\lambda_t=\lambda_\sigma-
\frac{(\lambda'_\sigma-\lambda_\sigma)
\langle\bm{r}'_\sigma-\bm{r}_\sigma,\bm{r}_\sigma\rangle}
{(1+\kappa)\|\bm{r}'_\sigma-\bm{r}_\sigma\|_2^2}$.
\STATE $\displaystyle
\bm{x}_{t-\Delta_t}=\xt-
\left[\textbf{f}(\xt,t)-\frac{1}{2}g^2(t)
\left(\s(\xt,t)+\lambda_t
\left(\frac{\alpha_t\ywave-\xt}{\sigma_t^2}
-\s(\xt,t)\right)\right)\right]\Delta_t$.
\ENDFOR
\end{algorithmic}
\end{algorithm}

\section{Related Work}
\label{sec:4}
\noindent\textbf{\textup{Diffusion Models and Bridges.}} Diffusion models have achieved state-of-the-art visual modeling capability. Based on these, various large-scale pre-trained diffusion models, such as Flux~\citep{flux2024,labs2025flux} and Wan~\citep{wan2025wan}, have emerged and demonstrated remarkable generative performance. Besides noise-to-visual generation, some works~\citep{zhou2023denoising,su2022dual,de2021diffusion,zheng2024diffusion} propose diffusion bridges to achieve the visual-to-visual translations. The typical paradigm is leveraging Doob's $h$-transform~\citep{doob1984classical,rogers2000diffusions} to link the paired samples and learn a joint distribution to transport between them. Our paper is inspired by the $h$-transform to achieve guided visual generation.

\noindent\textbf{\textup{Conditional Visual Generation.}} Conditional generation~\citep{zhan2024conditional,batzolis2021conditional,ho2022classifier} aims to handle various modality conditions (\eg, text, action, and visual samples) as guidance to constrain the generation. Beyond these common conditions, current research on conditional generation mainly focuses on visual condition signals (\eg, coarse sample, stroke, and pose map). Based on different types, the task can be specified as image restoration~\citep{wang2020deep,liang2021swinir,ren2023multiscale}, inverse problem~\citep{chung2022diffusion,kim2025flowdps,tarantola2005inverse,choi2021ilvr}, and conditional control~\citep{zhang2023adding,shen2024imagpose}, and so on. Meanwhile, the solutions also contain training-based~\citep{wang2022pretraining,murez2018image,parmar2024one} and inference-stage guidance~\citep{yu2023freedom,du2024demofusion,gao2025frequency,bu2026hiflow}. Our paper focuses on conditional generation w/ coarse samples as guidance and generates corresponding high-quality fine samples in a training-free manner.
\section{Experiments}
\label{sec:5}

\subsection{Coarse-Image Guided Generation}
\label{sec:5.1}
\noindent We first evaluated on image restoration tasks that aim to recover high-quality fine images from degraded samples. Specifically, this task entails recovering a clean image $\y$ given a coarse image $\ywave$.

\noindent{\textbf{\textup{Benchmark.}}} 
Following DPS~\citep{chung2022diffusion}, we use the validation set of FFHQ 256x256~\citep{karras2019style} to serve as the evaluation dataset. This dataset contains $1000$ clean images of human faces in $256*256$ resolution. For the evaluation metrics, we adopt Frechet Inception Distance~\citep{heusel2017gans} (\textit{FID}) and Learned Perceptual Image Patch Similarity~\citep{zhang2018unreasonable} (\textit{LPIPS}) to quantify the effectiveness. \textit{FID} can indicate the distribution similarity between synthetic images and ground-truth images, while \textit{LPIPS} focuses on image-level similarity.

\noindent{\textbf{\textup{Implementations.}}}
Following the protocol in DPS~\citep{chung2022diffusion}, we transform all clean images $\y$ by using a forward operator $\mathcal{A}$ and additive noise $\textbf{n}$ into coarse images $\ywave$, \ie, $\ywave = \mathcal{A}(\y) + \textbf{n}$. Depending on the forward operator type, it can be divided into four image restoration tasks: super-resolution (\textit{SR}), inpainting (\textit{Inpaint}), motion deblur (\textit{MD}), and Gaussian deblur (\textit{GD}). These coarse images $\ywave$ serve as guidance in the generation process. For fairness, we follow previous work~\citep{chung2022diffusion} and use the same pre-trained diffusion model from~\citep{dhariwal2021diffusion} and maintain other settings the same. During sampling, we set the initialized weight $\lambda_t=\sigma^5$.
\begin{table}[!t]
\centering
\setlength{\tabcolsep}{3pt}
\renewcommand{\arraystretch}{0.96}
\caption{Quantitative results of coarse-image guided generation on the FFHQ 256x256 validation dataset. \textbf{Bold}: best, \underline{underline}: second best.}
\vspace{-1em}
\label{tab:1}
\small
\resizebox{1\columnwidth}{!}{%
\begin{tabular}
{
    lccccccccc
}
\hline
\hline
\multirow{2}{*}{Method} & \multirow{2}{*}{\begin{tabular}[c]{@{}c@{}}Known\\ Operator\end{tabular}} & \multicolumn{2}{c}{SR} & \multicolumn{2}{c}{Inpaint} & \multicolumn{2}{c}{GD} & \multicolumn{2}{c}{MD} \\ 
\cmidrule(lr){3-4} \cmidrule(lr){5-6} \cmidrule(lr){7-8} \cmidrule(lr){9-10} 
& & {FID$_\downarrow$} & {LPIPS$_\downarrow$} & {FID$_\downarrow$} & {LPIPS$_\downarrow$} & {FID$_\downarrow$} & {LPIPS$_\downarrow$} & {FID$_\downarrow$} & {LPIPS$_\downarrow$} \\ 
\hline
ADMM-TV                 & $\checkmark$                                                  & 110.6            & 0.428            & 181.5          & 0.463         & 186.7            & 0.507            & 152.3           & 0.508           \\
ILVR~\citep{choi2021ilvr,song2020score}       &    $\checkmark$                                                                        & 96.72            & 0.563            & 76.54          & 0.612         & 109.0            & 0.403            & 292.2           & 0.657           \\
PnP-ADMM~\citep{chan2016plug}                & $\checkmark$                                                                  & 66.52            & 0.353            & 123.6          & 0.692         & 90.42            & 0.441            & 89.08           & 0.405           \\
MCG~\citep{chung2022improving}                     &  $\checkmark$                                                                   & 87.64            & 0.520            & \underline{29.26}          & \underline{0.286}         & 101.2            & 0.340            & 310.5           & 0.702           \\
DDRM~\citep{kawar2022denoising}                   &$\checkmark$                                                                  & 62.15            & 0.294            & 69.71          & 0.587         & 74.92            & 0.332            & -               & -               \\
DPS~\citep{chung2022diffusion}                   &   $\checkmark$                                                                  & 39.35            & \underline{0.214}            & \textbf{21.19}          & \textbf{0.212}         & 44.05            & 0.257            & 39.92           & 0.242           \\
SDEdit~\citep{meng2021sdedit}                  &                                           &    33.73           &    0.274     &      52.77           &    0.404            &    \textbf{35.30}              &    0.296              &    \textbf{34.41}             &   0.289               \\ 
BlindDPS~\citep{chihaoui2024blind}                  &                                           &      -         &    -     &     -            &      -          &  39.92                &     \textbf{0.216}             &    38.89             &    \textbf{0.204}              \\  
HiFlow~\citep{bu2026hiflow}                  &                                           &   \underline{33.48}            &   0.225      &     -            &      -          &  -                &     -             &    -             &    -              \\  
\textbf{Ours}                    &                                                                 &\cellcolor{mygray}\textbf{33.31}&\cellcolor{mygray}\textbf{0.213}&\cellcolor{mygray}{48.55}&\cellcolor{mygray}0.334&\cellcolor{mygray}\underline{38.17}&\cellcolor{mygray}\underline{0.253} &\cellcolor{mygray}\underline{35.23} &\cellcolor{mygray}\underline{0.237}
        \\

\hline\hline
\end{tabular}
}
\vspace{-1em}
\end{table}

\noindent{\textbf{\textup{Baselines.}}} 
We compared our method against: 1) six inverse-problem solutions (\cf, Fig.~\ref{fig:2}(b)): ADMM-TV, ILVR~\citep{choi2021ilvr,song2020score}, PnP-ADMM~\citep{chan2016plug}, MCG~\citep{chung2022improving}, DDRM~\citep{kawar2022denoising}, and DPS~\citep{chung2022diffusion}. These solutions require additional prior knowledge of the forward operator. We report their results from~\citep{chung2022diffusion}. 
2) The represented start-guided solution (\cf, Fig.~\ref{fig:2}(c)): SDEdit~\citep{meng2021sdedit}, which does not require knowing the forward operator. We set its $t_0=500$.
3) BlindDPS~\citep{chihaoui2024blind}, which relies on a trained blur kernel ($3$M steps) to mitigate the unknown operator problem.
4) HiFlow~\citep{bu2026hiflow}, an image super-resolution method.

\noindent{\textbf{\textup{Results.}}}
While the first $6$ baselines and BlindDPS require known forward operators or trained kernels, and HiFlow is designed for image super-resolution, our method generalizes across diverse tasks w/o needing task-specific operators or kernels. Despite its broad applicability, our approach remains highly competitive, ranking in the top two for $6/8$ metrics.
This verifies that our method provides a robust mechanism for coarse-image guidance, leading to structural preservation and fidelity.

\subsection{Coarse-Video Guided Generation}
\label{sec:5.2}

To further verify our effectiveness and generalization, we evaluated it on a camera-controlled video generation task. The camera-controlled video generation task is: given the first frame, corresponding camera intrinsics, and the pose, we aim to generate a video that follows a sequence of prescribed camera motions. To achieve this, we first use pretrained 3-D models to render a coarse video from the first frame and the prescribed camera motions. This coarse video is low-quality (warped) but can capture camera motion. Then it will serve as the guidance for the subsequent video generation.

\noindent{\textbf{\textup{Dataset.}}} 
Following previous camera-controlled generation work, we use DL3DV-10K~\citep{ling2024dl3dv} as our dataset for constructing coarse videos and evaluation. It contains annotated samples, each comprising a video with camera intrinsics and poses of all frames. We randomly select $160$ samples, retain the first $49$ frames, and resize them to $ 480p$. We use this subset as our dataset.

\noindent{\textbf{\textup{Implementations.}}}
We use DepthPro~\citep{bochkovskii2024depth} to extract the depth map of the first frame. Then project it into a 3-D point cloud. For the given sequence of camera motion poses, we render a sequence of corresponding images and concatenate them into the coarse video. The coarse video contains two parts. 1) Valid part: pixels that exist in the given first frame. 2) Invalid part: Pixels that do not exist in the given first frame. For the invalid part, the pixels are filled by nearest-neighbor assignment. All details about obtaining the coarse video are in the Appendix. During sampling, we only use the noise-level-aware initialization and drop the consistency adjustment for efficiency in video generation tasks, \ie, we set the weight function $\lambda_t=\sigma^\alpha$, and give these two parts different $\alpha$. For the valid part, $\alpha=4$, and for the invalid part, $\alpha=8$. For all hyperparameters, we followed the default settings of previous work~\citep{singer2025time,yang2024cogvideox}.

\begin{table}[!t]
\centering
\renewcommand{\arraystretch}{1}
\caption{Quantitative results of camera-controlled video generation on DL3DV.}
\vspace{-1em}
\label{tab:2}
\resizebox{0.83\columnwidth}{!}{%
\begin{tabularx}{\columnwidth}{l *{6}{>{\centering\arraybackslash}X}}
\hline
\hline
Method & MSE$_\downarrow$ & LPIPS$_\downarrow$ & FVD$_\downarrow$ & DINOv2$_\downarrow$& CLIP Cons.$_\uparrow$& Optical Flow$_\downarrow$\\ \hline
GT & -    &  -    &   -     &  -     &       0.974      &    -            \\\hline
Coarse Video & 11.46    &  0.276 & 15.55         &  0.220     &       0.973      &    41.3            \\
GWTF($\gamma=0.5$)   &  26.08   &    0.360     &  15.31         &   0.149    &  0.975           &   118.5            \\
GWTF($\gamma=0.7$) &  36.45    & 0.457    & 21.25    &      0.173    &   \textbf{0.984}          &   145.2            \\
TTM($t_w=4,t_s=8$)     & 23.50     & 0.382      & 15.69    &    0.147     &   0.980          &  157.2             \\
TTM($t_w=4,t_s=9$)     & 23.15     &  0.380     & 15.59    &      0.147    &      0.980       &     158.8          \\
\textbf{Ours}    & \cellcolor{mygray} \textbf{11.45}    &  \textbf{0.272}\cellcolor{mygray}    &  \cellcolor{mygray}  \textbf{13.26}    &  \cellcolor{mygray} \textbf{0.143}    &  \cellcolor{mygray}  0.972         &   \cellcolor{mygray} \textbf{38.7}           \\ \hline\hline
\end{tabularx}
}
\end{table}
\begin{figure*}[t]
    \centering
    \includegraphics[width=1\linewidth]{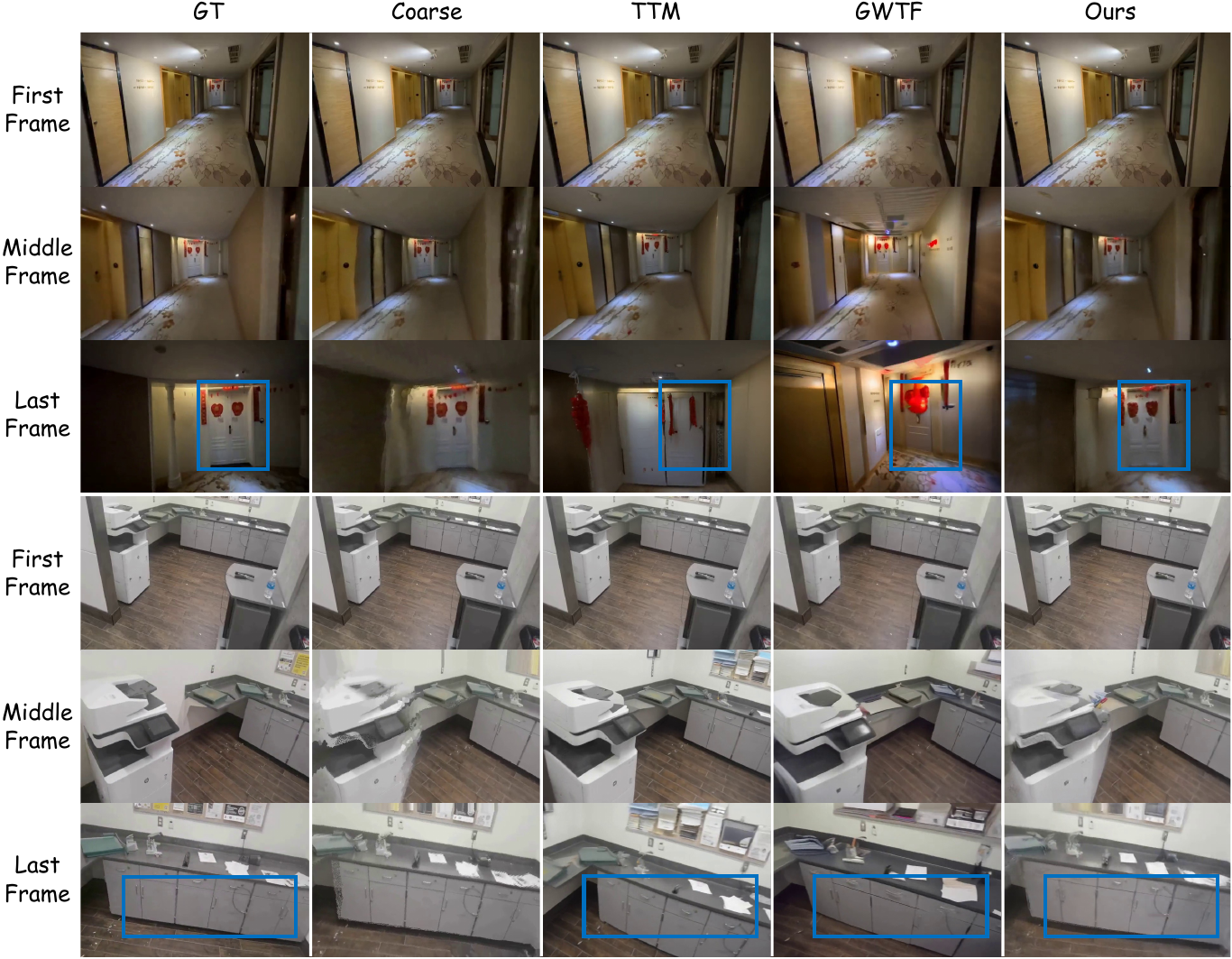}
    \vspace{-1em}
    \caption[]{\textbf{Qualitative comparisons on the subset of DL3DV-10K.} Our method shows better appearance alignment to the ground truth (see highlighted blue boxes).}
    \vspace{-1em}
    \label{fig:5}
\end{figure*}
\noindent{\textbf{\textup{Baselines.}}}
We compared our method with 1) Training-based GWTF~\citep{burgert2025go}: learning a noise warping based on optical flows to model the motion. We used the official trained model and set its key hyperparameter degradation value $\gamma$ to $0.5$ and $0.7$ as recommended in the original paper. 2) Training-free TTM~\citep{singer2025time}: leverages start-guided synthesis for two decoupled parts. This method also depends on a coarse video. For fairness, we use the same warped coarse as ours. We also took $t_{weak}=4$ ($t_w$), $t_{strong}=8$ ($t_s$), and $t_{weak}=4$, $t_{strong}=9$ according to the default and recommendations in their paper and code.

\noindent{\textbf{\textup{Metrics.}}}
We evaluated the quality of generated videos at the frame level using Mean Squared Error (\textit{MSE}) and \textit{LPIPS}~\citep{zhang2018unreasonable} relative to the ground-truth videos. Additionally, we calculated the distributional difference between the ground-truth videos and the synthetic videos using Frechet Video Distance~\citep{unterthiner2018towards} (\textit{FVD}). We also calculated the average semantic similarity of each frame between ground-truth videos and the synthetic videos with DINOv2 Distance~\citep{oquab2023dinov2} (\textit{DINOv2}). We used the CLIP similarity~\citep{radford2021learning} between two consistent frames (\textit{CLIP Cons.}) to evaluate the temporal consistency.
For camera motion consistency, we calculated the Mean Squared Error between the RAFT-estimated optical flows~\citep{teed2020raft} of the ground-truth videos and those of the synthetic videos (\textit{Optical Flow}).

\noindent{\textbf{\textup{Results.}}}
As shown in Table~\ref{tab:2}, our method achieves the best performance across all metrics except \textit{CLIP Cons.}. This indicates that our synthetic videos are closest to ground-truth videos and have the best motion consistency. For the \textit{CLIP Cons.}, although our method has the lowest $0.972$, we should note that the ground truth only achieves $0.974$ on this metric. This indicates that other methods even outperform the ground truth on this metric by sacrificing motion consistency to get quite similar consecutive frames. Additionally, we provide qualitative comparisons in Figure~\ref{fig:5}.

\subsection{Ablations of the Weight Function $\lambda$}
\label{sec:5.3}
\begin{wraptable}{R}{0.44\columnwidth}
\centering
\vspace{-2em}
\setlength{\tabcolsep}{5pt}
\renewcommand{\arraystretch}{0.92}
\small
\resizebox{\linewidth}{!}{%
\begin{tabular}{lcccc}
\hline
\hline
\multicolumn{1}{c}{$\lambda$} & SR & Inpaint & GD & MD \\ 
\hline
$0.0$                                                              &    0.586         &   0.586       &    0.586         &   0.586         \\
$0.5$                                                              &   0.456          &    0.989      &     0.445        & 0.491           \\
$1.0$                                                              &   0.472          &   1.008       &      0.483       &  0.546          \\
$2.0$                                                              &   0.466          &    1.009      &   0.486          &    0.552        \\
$\sin(\frac{\pi t}{2T})$                                           &    0.224         &   0.448       &    0.293         &    0.344        \\
$\frac{t}{T}$                                                      &   0.223          &   0.406       &     0.274        &     0.344       \\
$(\frac{t}{T})^3$                                                  &    0.380         &    0.369      &     0.393        &    0.428        \\
$(\frac{t}{T})^5$                                                  &    0.466         &    0.461      &       0.472      &    0.480        \\
$\sigma$                                                           &    0.268         &    0.752      &     0.359        &   0.412         \\
$\sigma^3$                                                         &   \textbf{0.208} &   0.433       &    0.257         &   0.340 \\
$\sigma^5$                                                         &    0.229         &     0.341     &  0.260           &       0.247     \\
\textbf{ours} ($\sigma^5$)                                                         &  0.213           &   \textbf{0.334}       &     \textbf{0.253}        &   \textbf{0.237}         \\
\hline\hline
\end{tabular}
}
\vspace{-1em}
\caption{The ablation of weight schedule $\lambda$.}
\vspace{-2em}
\label{tab:3}
\end{wraptable}
Table~\ref{tab:3} compares constant, sinusoidal, timestep-based, and noise-level-aware designs of $\lambda$ across four image restoration tasks using \textit{LPIPS} (lower is better). Among the non-adaptive designs, schedules based on powers of $\sigma$ consistently yield the best performance. This agrees with the error analysis in Section~\ref{sec:3.3}: the approximation error increases as the noise level decreases, motivating a schedule that progressively reduces the influence of the coarse approximation. These results distinguish our principled, noise-level-aware weighting from naive score blending and highlight its role in achieving high-quality guided synthesis.
Adding the consistency-based correction further improves performance and yields the best overall results, demonstrating the effectiveness of our error control guided by both factors during sampling.

\subsection{Additional Visual Generation Tasks}
\label{sec:6.4}


\begin{wraptable}{R}{0.72\columnwidth}
\centering
\vspace{-1em}
\setlength{\tabcolsep}{3pt}
\renewcommand{\arraystretch}{1}
\small
\resizebox{\linewidth}{!}{%
\begin{tabular}{lccccccc}
\hline\hline
\multirow{2}{*}{Method} & \multicolumn{5}{c}{Source Consistency} & \multicolumn{2}{c}{Semantic Alignment} \\
\cline{2-8}
 & Distance$_\downarrow$ & PSNR$_\uparrow$ & LPIPS$_\downarrow$ & MSE$_\downarrow$  & SSIM$_\uparrow$ & CLIP Entire$_\uparrow$ & CLIP Edited$_\uparrow$\\ \hline
ODE-Inv      &  0.074    & 17.57 & 0.240 & 0.024 & 0.691 &  24.57     &   21.73     \\
SDEdit     & 0.036    & 22.57 & 0.119 & 0.008 & 0.747 & 24.56        & 21.95        \\
iRFDS     & 0.069    & 18.81 & 0.191 &0.021  & 0.738 &   25.12      &   21.95      \\
FlowEdit  & 0.036    & 23.02 & \underline{0.082} & 0.007 & \underline{0.842} & \textbf{25.98}        & \textbf{22.81}        \\
FlowAlign & \underline{0.028}    & \underline{25.50} &     \textbf{0.053} & \textbf{0.004} & \textbf{0.879} & 25.28        & 22.00        \\
 Ours &\textbf{0.017}&\textbf{25.98}&0.097&\textbf{0.004}&0.819&\underline{25.41}& \underline{22.37} \\

 \hline\hline
\end{tabular}%
}
\vspace{-1em}
\caption{Quantitative comparisons with five baselines for image editing.}
\vspace{-1em}
\label{tab:4}
\end{wraptable}
\textbf{Image Editing}.
To further explore the generalization ability of our method, we conducted experiments on the image editing task, \ie, given a source image, a source prompt, and a target prompt, we edit the source image into a target image. Specifically, our method uses the source image as the $\ywave$, and $\s$ is the predicted score by a pretrained text-to-image models that adopt the target prompt as the condition. We set the sampling timestep to $50$, the target guidance scale to $13.5$, and $\lambda_{t}=\sigma_t$.
We compared our method with five state-of-the-art flow-based editing methods that were implemented with SD-3-medium~\citep{esser2024scaling}: \textit{ODE inversion}, \textit{SDEdit}~\citep{meng2021sdedit}, \textit{iRFDS}~\citep{yang2024text}, \textit{FlowEdit}~\citep{kulikov2024flowedit} and \textit{FlowAlign}~\citep{kim2025flowalign}. The results for these baselines are from \textit{FlowAlign} on PIE-Bench~\citep{ju2023pnp} and we report the seven common metrics to evaluate both source consistency (\textit{Distance}, \textit{PSNR}, \textit{LPIPS}, \textit{MSE}, \textit{SSIM}) and Semantic Alignment (\textit{Entire CLIP} and \textit{Edited CLIP Score}). The results are shown in Tab.~\ref{tab:4}.


\begin{wrapfigure}{r}{0.5\textwidth}
    \centering
    \vspace{-1em}
    \includegraphics[width=0.48\textwidth]{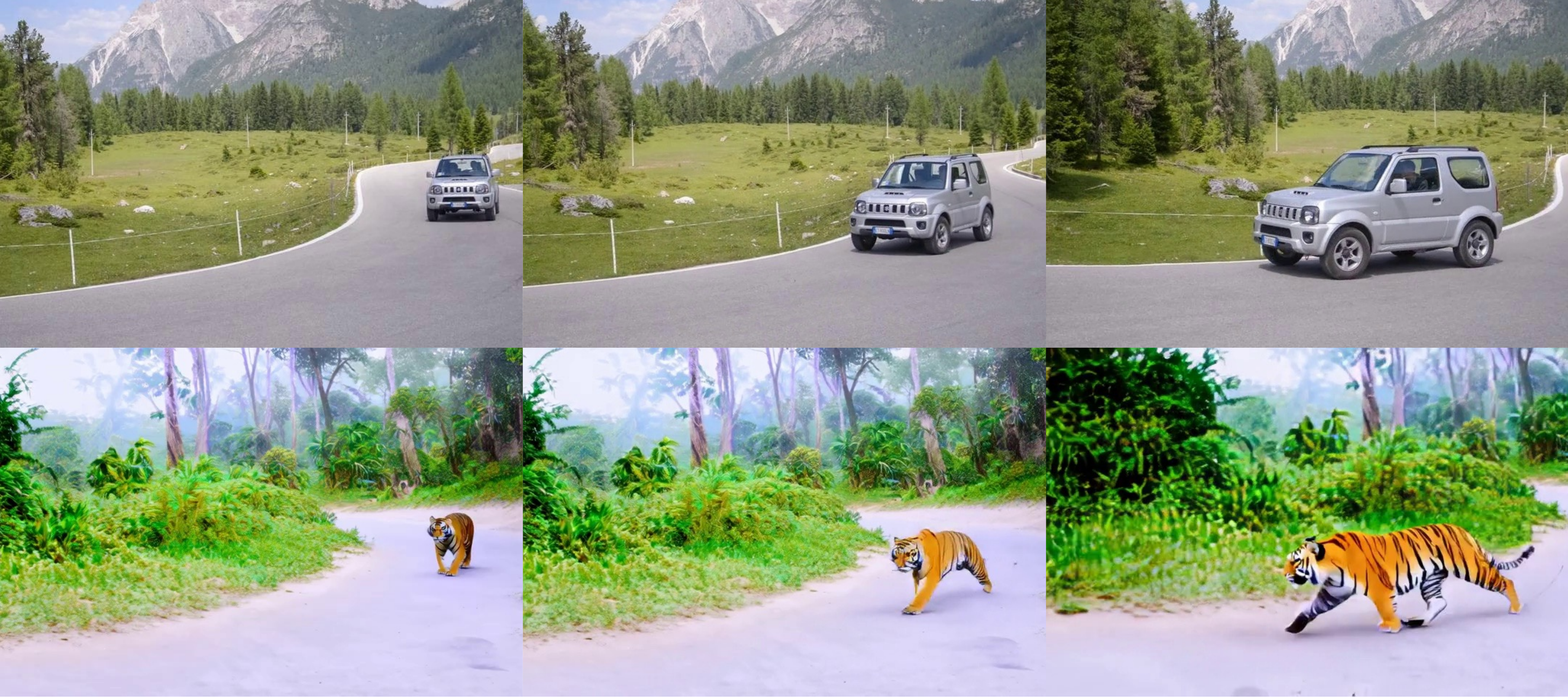}
    \vspace{-1em}
    \caption{Generalize to video motion transfer.}
    \vspace{-2em}
    \label{fig:4}
\end{wrapfigure}
\textbf{Video Motion Transfer}. 
We further apply our method to video motion transfer, translating the temporal dynamics of a source video into a new semantic context. By using the source video (\eg, a car driving in a field) as the coarse guidance $\ywave$ and a target text prompt (\eg, ``a tiger walking in a forest") to dictate the semantic score $\s$, our framework seamlessly integrates both signals. As illustrated in Figure~\ref{fig:4}, the generated video successfully preserves the original motion trajectories while synthesizing high-fidelity target semantics, demonstrating the potential applicability of our method across diverse modalities.

\section{Conclusion}
\label{sec:6}

We proposed \textbf{Adaptive Weighted $h$-Transform Sampling} for coarse-guided visual generation. 
We inject the guidance during the sampling process by designing an $h$ function that can approximately drive the generation towards the underlying ideal sample. We further design a weight scheduler to restrict the approximation error for better synthetic results. Extensive ablations show the generalization and effectiveness. In the future, probability transition modification mechanisms (represented by $h$-transform) can be further explored in more general conditional generation applications.



\bibliography{iclr2027_conference}
\bibliographystyle{iclr2027_conference}
\renewcommand{\thesection}{A.\arabic{section}}
\renewcommand*{\thesubsection}{A.\arabic{subsection}}
\renewcommand{\theequation}{A\arabic{equation}}
\renewcommand{\thetable}{A\arabic{table}}
\renewcommand{\thefigure}{A\arabic{figure}}

\newpage
\appendix
\noindent\textbf{\Large Appendix}
\vspace{0.1in}

This appendix is organized as follows:

\begin{itemize}[leftmargin=*]
\itemsep-0.2em
\item Section~\ref{sec:a.1} gives the proof of equivalent marginal distributions between the reversed-SDE and its corresponding PF-ODE.
\item Section~\ref{sec:a.2} demonstrates the detailed derivation of $\happ$.
\item Section~\ref{sec:a.3} provides the Weighted $h$-Transform Inference equation for velocity-based Optimal Transport Flow Matching (OT-FM) and noise-based Variance-Preserving SDE (VP-SDE).
\item Section~\ref{sec:a.4} indicates the implementation details of our method, including how we obtain the coarse videos, text prompts of video generation, and how we set related hyperparameters.
\item Section~\ref{sec:a.5} provides additional qualitative results.  
\end{itemize}

\newpage
\subsection{Equivalence Proof of Marginal Distributions}
\label{sec:a.1}
In this section, we provide the formal proof that the stochastic differential equation (SDE) in Eq.~(\ref{eq:6}) and the ordinary differential equation (ODE) in Eq.~(\ref{eq:7}) share the same marginal distribution.

Let $q_t(\xt)$ denote the marginal probability density of the latent $\xt$ at time $t$.
We first write the drift term of the SDE (Eq.~(\ref{eq:6})) as:
\begin{equation}
    \bm{\mu}_{\text{SDE}}(\xt, t) = \mathbf{f}(\xt, t) - g^2(t)\grad(\xt) - g^2(t)\hgt
\end{equation}
The evolution of the probability density $q_t(\xt)$ for this reverse-time SDE is governed by the Fokker-Planck equation~\citep{maoutsa2020interacting} (F-P equation). Taking into account the negative time increment in the reverse process, the corresponding F-P equation is given by:
\begin{equation}
    \frac{\partial q_t}{\partial t} = -\nabla \cdot (\bm{\mu}_{\text{SDE}} q_t) - \frac{1}{2} g^2(t) \Delta q_t
    \label{eq:fpe_initial}
\end{equation}

Then, we substitute the mathematical identity $\Delta q_t = \nabla \cdot (\nabla q_t) = \nabla \cdot (q_t \nabla \log q_t)$ into Eq.~(\ref{eq:fpe_initial}) yields:
\begin{equation}
    \frac{\partial q_t}{\partial t} = -\nabla \cdot (\bm{\mu}_{\text{SDE}} q_t) - \frac{1}{2} g^2(t) \nabla \cdot (q_t \nabla \log q_t)
\end{equation}
We can reformulate the equation as a standard continuity equation:
\begin{equation}
    \frac{\partial q_t}{\partial t} = -\nabla \cdot \left( \underbrace{\left[ \bm{\mu}_{\text{SDE}} + \frac{1}{2} g^2(t) \nabla \log q_t \right]}_{\bm{v}_{\text{ODE}}} q_t \right)
    \label{eq:continuity_eq}
\end{equation}

Eq.~(\ref{eq:continuity_eq}) takes the form $\frac{\partial q_t}{\partial t} = -\nabla \cdot (\bm{v}_{\text{ODE}} q_t)$, which precisely describes the density evolution of a deterministic ODE $\mathrm{d}\bm{x} = \bm{v}_{\text{ODE}}\mathrm{d}t$. Therefore, the equivalent deterministic drift $\bm{v}_{\text{ODE}}$ that shares the same marginal density $q_t$ is:
\begin{equation}
    \bm{v}_{\text{ODE}} = \bm{\mu}_{\text{SDE}} + \frac{1}{2} g^2(t) \nabla \log q_t(\xt)
    \label{eq:v_ode_intermediate}
\end{equation}

In the context of diffusion models, the goal of introducing the additional term $\hgt$ is to shift the unconditional score to the target conditional score. By definition, the exact marginal score of the process is structurally related to the unconditional score via:
\begin{equation}
    \nabla \log q_t(\xt) = \grad(\xt) + \hgt
    \label{eq:score_identity}
\end{equation}

Finally, we substitute the definition of $\bm{\mu}_{\text{SDE}}$ and the score identity from Eq.~(\ref{eq:score_identity}) into Eq.~(\ref{eq:v_ode_intermediate}):
\begin{align}
    \bm{v}_{\text{ODE}} &= \left[ \mathbf{f}(\xt, t) - g^2(t)\grad(\xt) - g^2(t)\hgt \right] \nonumber \\
    &\quad + \frac{1}{2} g^2(t) \left( \grad(\xt) + \hgt \right)
\end{align}
Simplifying the terms, we obtain:
\begin{equation}
    \bm{v}_{\text{ODE}} = \mathbf{f}(\xt, t) - \frac{1}{2} g^2(t)\left( \grad(\xt) + \hgt \right)
\end{equation}

This derived drift $\bm{v}_{\text{ODE}}$ is identical to the drift of the ODE presented in Eq.~(\ref{eq:7}). Since the same continuity equation governs the probability density of the given SDE and this resulting ODE, we conclude that they exhibit the same marginal distribution $q_t(\xt)$ at all times $t$.

\newpage
\subsection{Derivation of \texorpdfstring{$\happ$}{happ}}
\label{sec:a.2}
In this section, we provide the detailed derivation of $\happ$.

Based on Eq.~(\ref{eq:10}), we can get the probability density function of $p_t(\xt|\xzero = \ywave)$:
\begin{equation}
\begin{aligned}
p_t(\xt|\xzero = \ywave) = \frac{1}{(2\pi\sigma_t^2)^{d/2}} \exp\left( -\frac{\|\xt - \alpha_t\ywave\|^2}{2\sigma_t^2} \right).
\end{aligned}
\end{equation}

Then take the $\log$ of both sides of the equation and calculate the gradient:
\begin{equation}
\begin{aligned}
\grad(\xt|\xzero = \ywave) &= \nabla_{\xt} \left[ -\frac{d}{2} \log(2\pi \sigma_t^2) - \frac{\|\xt - \alpha_t \ywave\|^2}{2\sigma_t^2} \right] \\
&= -\frac{1}{2\sigma_t^2} \nabla_{\xt} \|\xt - \alpha_t \ywave\|^2 \\
&= -\frac{2(\xt - \alpha_t \ywave)}{2\sigma_t^2} \\
&= -\frac{\xt - \alpha_t \ywave}{\sigma_t^2} = \frac{1}{\sigma_t^2}(\alpha_t\ywave-\xt).
\end{aligned}
\end{equation}
Then we substitute it into Eq.~(\ref{eq:9}) and get the final $\happ$, \ie, Eq.~(\ref{eq:12}).

\newpage
\subsection{Weighted $h$-Transform Inference for Various SDEs}
\label{sec:a.3}

We derived our method, Weighted $h$-Transform Inference, from the Score Matching perspective in our paper, and we claim that our method is compatible with other diffusion models like DDPM and Flow Matching. In this section, we provide the derivation of the inference equation of $v$-based Optimal Transport Flow Matching (OT-FM) and $\epsilon$-based Variance-Preserving SDE (VP-SDE).

\subsubsection{OT-FM with $v$-prediction.}
In Flow Matching, the velocity field $v_\theta(\xt, t)$ corresponds to the PF-ODE drift. Thus, we have the relation:
\begin{equation}
    v_\theta = \mathbf{f}(\xt, t) - \frac{1}{2}g^2(t) \s \quad \implies \quad \frac{1}{2}g^2(t) \s = \mathbf{f}(\xt, t) - v_\theta
\end{equation}

Substituting this relation into the ODE (Eq.~(\ref{eq:15})), we rewrite the drift in terms of $v_\theta$:
\begin{align}
    \mathrm{d}\mathbf{x} &= \left[ v_\theta + \lambda_\sigma \left( \frac{g^2(t)}{2\sigma_t^2}(\xt - \alpha_t \ywave) + \frac{1}{2}g^2(t)\s \right) \right]\mathrm{d}t \nonumber \\
    &= \left[ v_\theta + \lambda_\sigma \left( \frac{g^2(t)}{2\sigma_t^2}(\xt - \alpha_t \ywave) + \mathbf{f}(\xt, t) - v_\theta \right) \right]\mathrm{d}t \label{eq:intermediate_ode}
\end{align}

For OT-FM, the trajectory is defined by $\sigma_t = 1 - \alpha_t$. Under a linear schedule where $\alpha_t = 1 - t$ and $\sigma_t = t$, we have $\dot{\alpha}_t = -1$ and $\dot{\sigma}_t = 1$. The corresponding drift and squared diffusion coefficients are:

\begin{equation}
\label{eq:f}
        \mathbf{f}(\xt, t) = \frac{\dot{\alpha}_t}{\alpha_t} \xt
\end{equation}

\begin{equation}
\label{eq:g}
     g^2(t) = 2\sigma_t\dot{\sigma}_t - 2\frac{\dot{\alpha}_t}{\alpha_t}\sigma_t^2 = -2\dot{\alpha}_t\sigma_t \left( 1 + \frac{\sigma_t}{\alpha_t} \right) = -2\dot{\alpha}_t \frac{\sigma_t}{\alpha_t}
\end{equation}

We now simplify the guidance term from Eq.~(\ref{eq:intermediate_ode}) using these coefficients:
\begin{align}
    \frac{g^2(t)}{2\sigma_t^2}(\xt - \alpha_t \ywave) + \mathbf{f}(\xt, t) 
    &= \frac{-2\dot{\alpha}_t \frac{\sigma_t}{\alpha_t}}{2\sigma_t^2} (\xt - \alpha_t \ywave) + \frac{\dot{\alpha}_t}{\alpha_t} \xt \nonumber \\
    &= -\frac{\dot{\alpha}_t}{\alpha_t\sigma_t} (\xt - \alpha_t \ywave) + \frac{\dot{\alpha}_t}{\alpha_t} \xt \nonumber \\
    &= \xt \dot{\alpha}_t \left( \frac{1}{\alpha_t} - \frac{1}{\alpha_t\sigma_t} \right) + \frac{\dot{\alpha}_t}{\sigma_t} \ywave
\end{align}

Using the OT constraint $\sigma_t - 1 = -\alpha_t$, the term simplifies perfectly:
\begin{equation}
    \xt \dot{\alpha}_t \left( \frac{\sigma_t - 1}{\alpha_t\sigma_t} \right) + \frac{\dot{\alpha}_t}{\sigma_t} \ywave 
    = -\frac{\dot{\alpha}_t}{\sigma_t} \xt + \frac{\dot{\alpha}_t}{\sigma_t} \ywave 
    = -\dot{\alpha}_t \left( \frac{\xt - \ywave}{\sigma_t} \right)
\end{equation}

Since $\dot{\alpha}_t = -1$, this term evaluates exactly to $\frac{\xt - \ywave}{\sigma_t}$. Substituting this back into Eq.~(\ref{eq:intermediate_ode}) yields the final velocity-based guided ODE:
\begin{equation}
    \mathrm{d}\mathbf{x} = \left[ v_\theta + \lambda_\sigma \left( \frac{\xt - \ywave}{\sigma_t} - v_\theta \right) \right]\mathrm{d}t
\end{equation}

\subsubsection{VP-SDE with $\epsilon$-Prediction.}
For a VP forward process $\xt = \alpha_t \xzero + \sigma_t \boldsymbol{\epsilon}$ constrained by $\alpha_t^2 + \sigma_t^2 = 1$, the drift and diffusion coefficients are also uniquely determined by the schedule $\alpha_t$, \ie, Eq.~(\ref{eq:f}) and Eq.~(\ref{eq:g}).

In the $\epsilon$-prediction framework, the score function is parameterized as $\s = -\frac{\epsilon_\theta}{\sigma_t}$. Substituting $\mathbf{f}$, $g^2(t)$, and $\s$ into Eq.~(\ref{eq:15}), we obtain:
\begin{align}
    \mathrm{d}\mathbf{x} &= \left[ \frac{\dot{\alpha}_t}{\alpha_t}\xt - \frac{1}{2}\left(-2\frac{\dot{\alpha}_t}{\alpha_t}\right) \left( -\frac{\epsilon_\theta}{\sigma_t} + \lambda_\sigma \left(\frac{\alpha_t \ywave - \xt}{\sigma_t^2} + \frac{\epsilon_\theta}{\sigma_t}\right)\right) \right]\mathrm{d}t \nonumber \\
    &= \frac{\dot{\alpha}_t}{\alpha_t} \left[ \xt - \frac{1}{\sigma_t}\left( \epsilon_\theta - \lambda_\sigma \left( \frac{\alpha_t \ywave - \xt}{\sigma_t} + \epsilon_\theta \right) \right) \right]\mathrm{d}t \label{eq:vp_alpha_sigma_intermediate}
\end{align}

We define the pseudo-target noise corresponding to the condition $\ywave$ as:
\begin{equation}
    \ewave = \frac{\xt - \alpha_t \ywave}{\sigma_t}
\end{equation}

By observing that $\frac{\alpha_t \ywave - \xt}{\sigma_t} = -\ewave$, we can substitute this into the inner guidance term in Eq.~(\ref{eq:vp_alpha_sigma_intermediate}):
\begin{equation}
    \epsilon_\theta - \lambda_\sigma \left( -\ewave + \epsilon_\theta \right) = \epsilon_\theta + \lambda_\sigma \left( \ewave - \epsilon_\theta \right)
\end{equation}

This reveals the effective guided noise prediction $\hat{\epsilon}_\theta$:
\begin{equation}
    \hat{\epsilon}_\theta = \epsilon_\theta + \lambda_\sigma \left( \ewave - \epsilon_\theta \right)
\end{equation}

Consequently, the guided PF-ODE simplifies elegantly to the standard unguided formulation, driven entirely by the linearly interpolated noise $\hat{\epsilon}_\theta$:
\begin{equation}
    \mathrm{d}\mathbf{x} = \frac{\dot{\alpha}_t}{\alpha_t} \left( \xt - \frac{\hat{\epsilon}_\theta}{\sigma_t} \right) \mathrm{d}t
\end{equation}

\newpage
\subsection{Implementation details}
\label{sec:a.4}

\subsubsection{Obtaining the Coarse Videos} In Section~\ref{sec:5.2}, we use 3-D models to render an image into a corresponding coarse warped video based on its camera intrinsics and poses. First, we leverage a pretrained depth estimation model, DepthPro~\citep{bochkovskii2024depth}, to get the depth map of the first frame image. Then we reproject this depth map into a point cloud and render a sequence of images for given camera motion poses. To ensure motion quality, we filter out static or minimal-movement sequences where the camera translation extent is less than $8$. 

To resolve the scale ambiguity of monocular depth maps, we align them with the camera trajectories using SIFT feature matching~\citep{lowe2004distinctive} between the reference frame and subsequent frames (sampled at $5$-frame intervals). Matches are refined using a $k$-nearest neighbor search ($k=2$) and a Lowe's ratio test threshold of $0.75$. After computing candidate scales via projection constraints, we discard the top and bottom $5\%$ outliers and apply the median value as the final scale. The reference image is warped to target views by unprojecting valid pixels (depth $>$ $0$) into 3-D space using the inverse intrinsic matrix and the calibrated depth. These points are transformed into the target camera coordinate system and splatted onto the 2-D image plane. 

To correctly resolve self-occlusions during rendering, we implement a $Z$-buffering approach by sorting the projected points in descending order of depth. To address deocclusion holes inherently caused by forward splatting, we implement a global nearest-neighbor filling strategy utilizing a k-d tree built on valid pixel coordinates. The corresponding visibility masks are further refined using morphological opening with a $5*5$ rectangular kernel to eliminate isolated projection noise and smooth edges. More details can be found in the code.

\subsubsection{Image-to-Video Prompts Preparation.} For prompts generation, we follow the protocol of CogVideoX~\citep{yang2024cogvideox}, \ie, leverage the pretrained Vision-Language model GPT-4o~\citep{hurst2024gpt} to generate caption for the given first frame. These generated prompts will be used as the text condition for subsequent image-to-video generation. 

\subsubsection{The Implementation of Using Wan2.2 for Camera-controlled Video Generation.}

For Wan2.2~\citep{wan2025wan}, we change the length and resolution of evaluated videos, \ie, the frame number from $49$ to $81$ and the resolution from $720\times480$ to $832\times480$. During sampling, we set the weight function $\lambda_\sigma=\sigma^\alpha$ and give these two parts different $\alpha$. For the valid part, $\alpha=16$, and for the invalid part, $\alpha=20$.
For other hyperparameters of Wan2.2, we follow the implementations of TTM~\citep{singer2025time}.

\newpage
\subsection{Additional results}
\label{sec:a.5}
\subsection{Compatibility with Flow Matching Models}
\begin{wrapfigure}{r}{0.6\linewidth}
\centering
\vspace{-2em}
\includegraphics[width=1\linewidth]{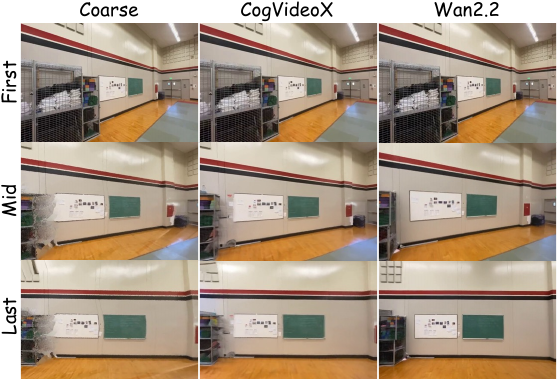}
\vspace{-2em}
\caption{Our method is compatible with both score-based CogVideoX and Flow-based Wan2.2.}
\label{fig:7}

\end{wrapfigure}
To further verify the generalization ability of our method, we evaluated it on the camera-controlled video generation task (\cf, Section~\ref{sec:5.2}) by using a flow-based model, Wan2.2~\citep{wan2025wan}. We gave qualitative results of Wan2.2 in Figure~\ref{fig:7}. We can observe that when the pretrained model is Wan2.2, the synthetic videos also yield a decent synthetic result (even better than CogVideoX since Wan2.2 is a stronger base model) under the guidance of the coarse video. This indicates that our method is compatible with both score-based models like CogVideoX and flow-based models like Wan2.2, showing good generalization. 
This indicates that, although our proposed Weighted $h$-Transform Sampling is based on the score perspective, it can generally be compatible with flow matching and score matching.

\newpage
\subsubsection{Additional Qualitative Results}
\begin{figure*}[!t]
    \centering
    \includegraphics[width=1\linewidth]{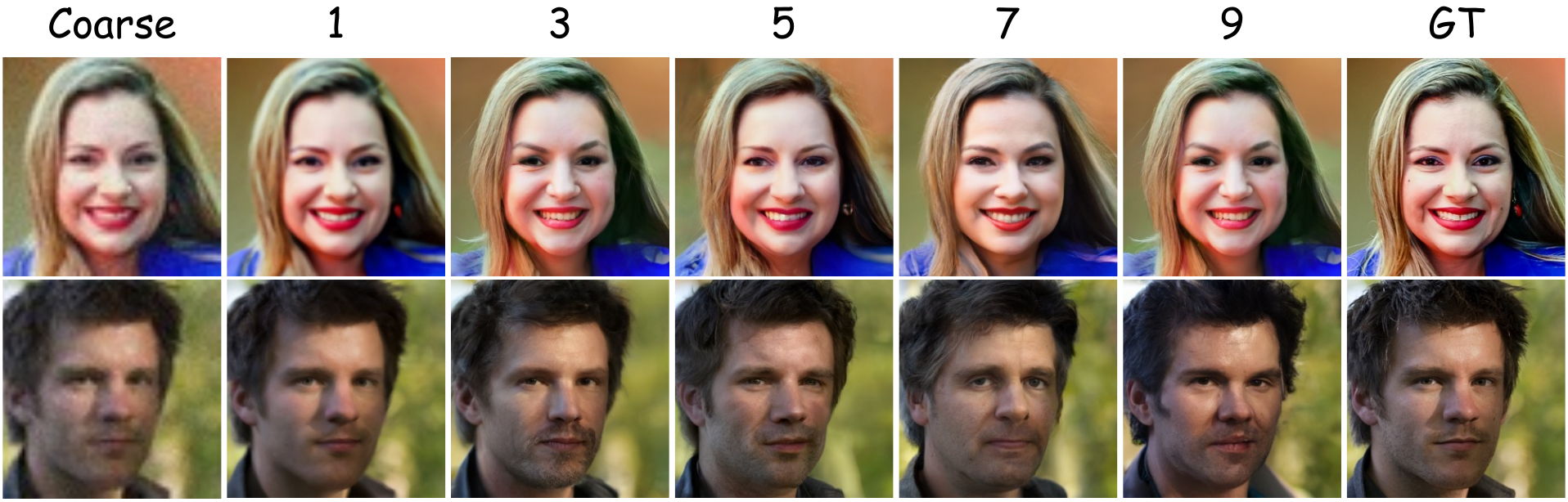}
    \caption[]{\textbf{Super-resolution results by using different $\alpha$}. A large $\alpha$ leads to a small weight, and vice versa (weight scheduler $\lambda_\sigma=\sigma_t^\alpha$).}
    \label{fig:8}
\end{figure*}

\begin{figure*}[!t]
    \centering
    \includegraphics[width=1\linewidth]{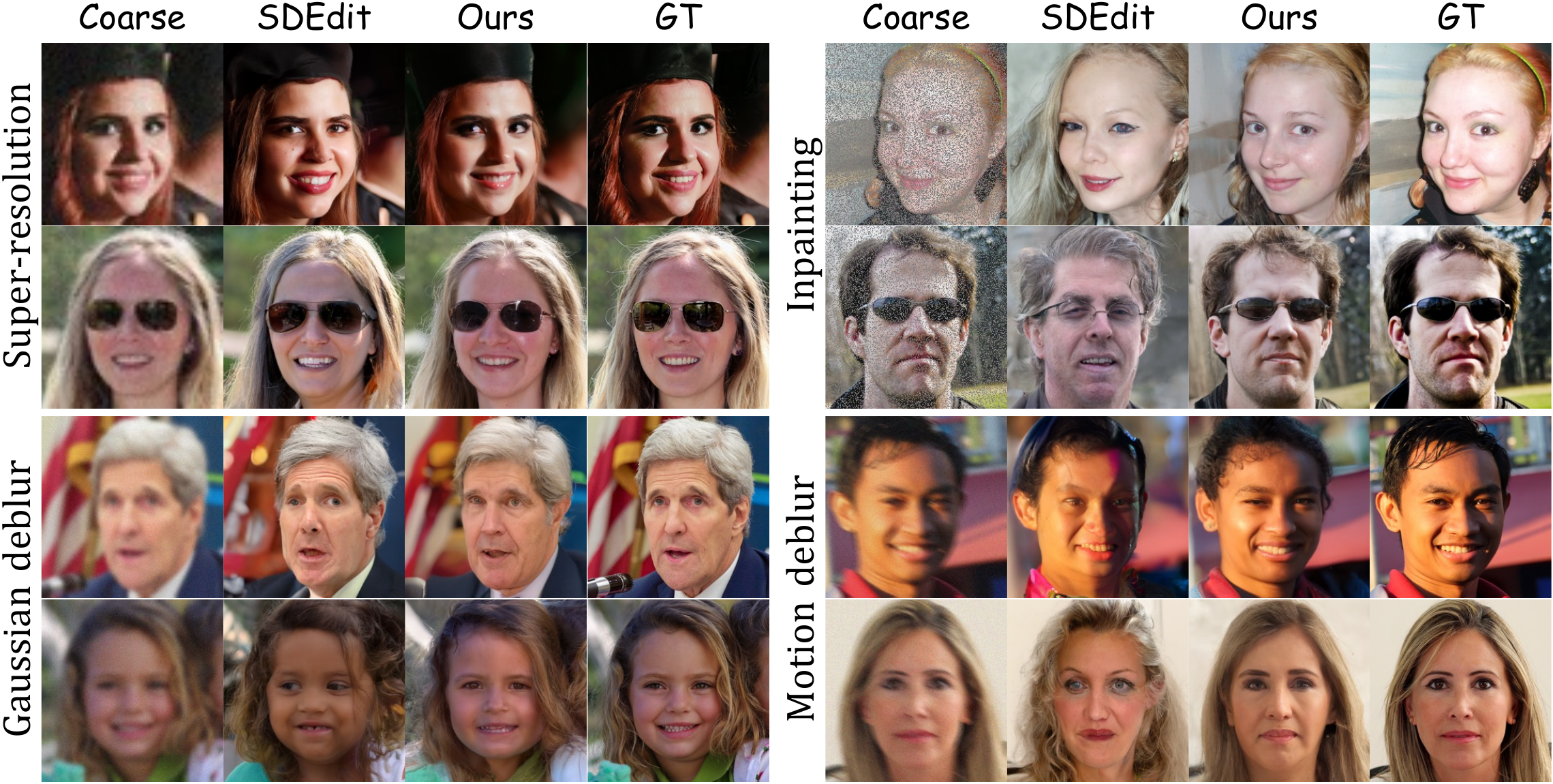}
        \caption{\textbf{Additional qualitative results on image restoration tasks.}}
    \label{fig:6}
\end{figure*}

\begin{figure*}[!t]
    \centering
    \includegraphics[width=0.65\linewidth]{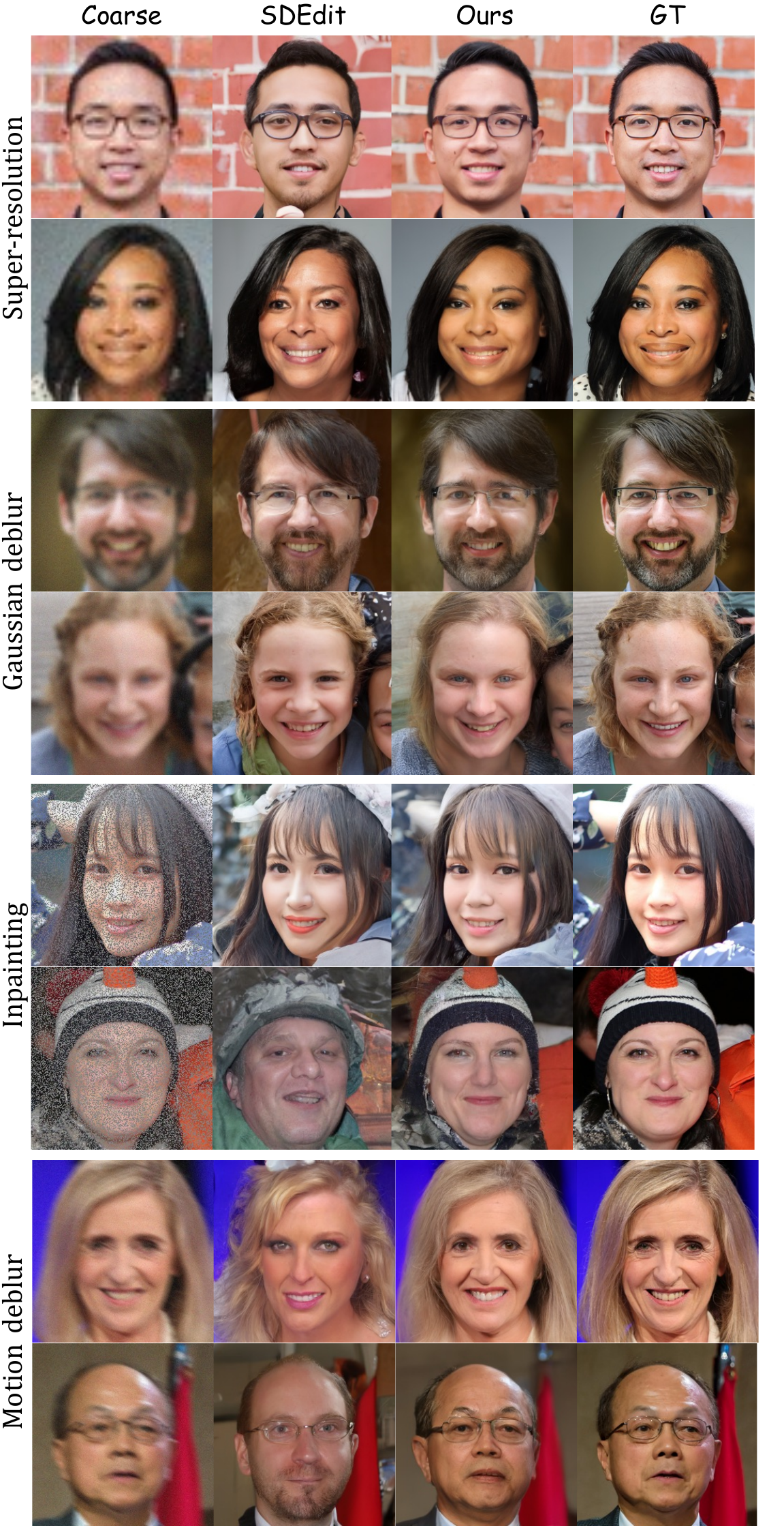}
    \caption[]{Additional qualitative comparisons with SDEdit on image restoration tasks.}
    \label{fig:13}
\end{figure*}

\begin{figure*}[!t]
    \centering
    \includegraphics[width=1\linewidth]{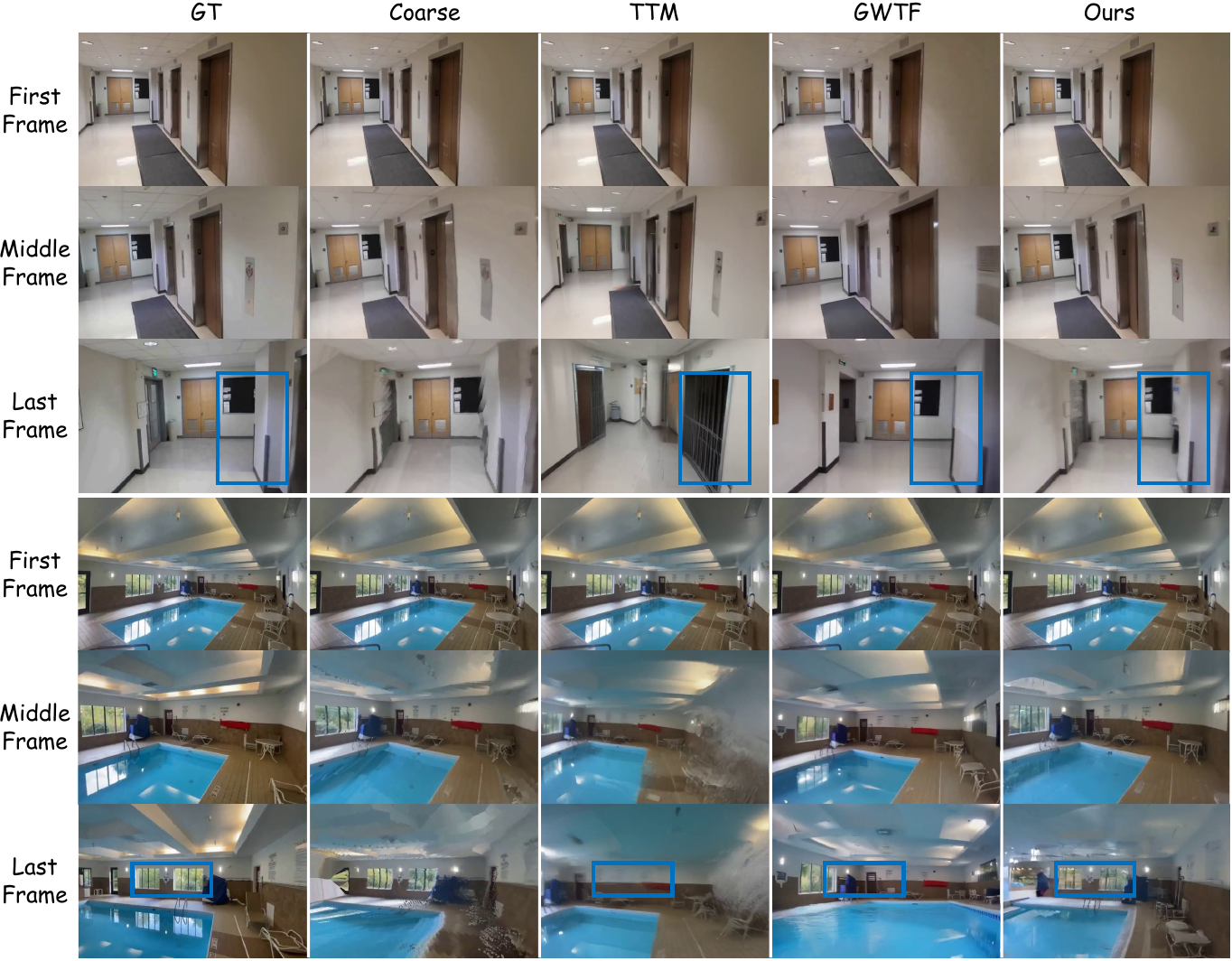}
    \caption[]{Additional camera-controlled video generation qualitative comparisons.}
    \label{fig:11}
\end{figure*}

\begin{figure*}[!t]
    \centering
    \includegraphics[width=1\linewidth]{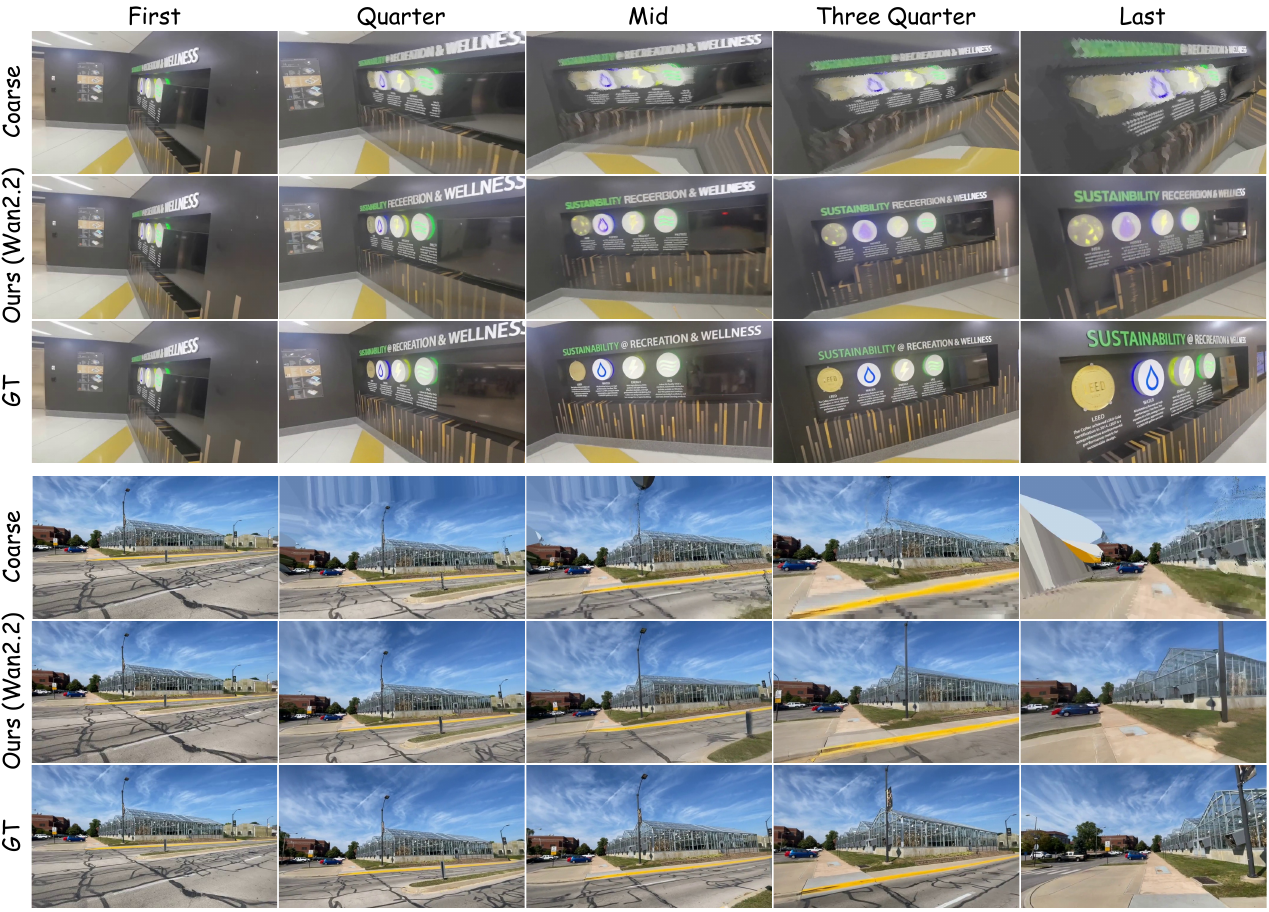}
    \caption[]{Additional camera-controlled video generation results using Wan2.2.}
    \label{fig:12}
\end{figure*}

\begin{figure*}[!t]
    \centering
    \includegraphics[width=1\linewidth]{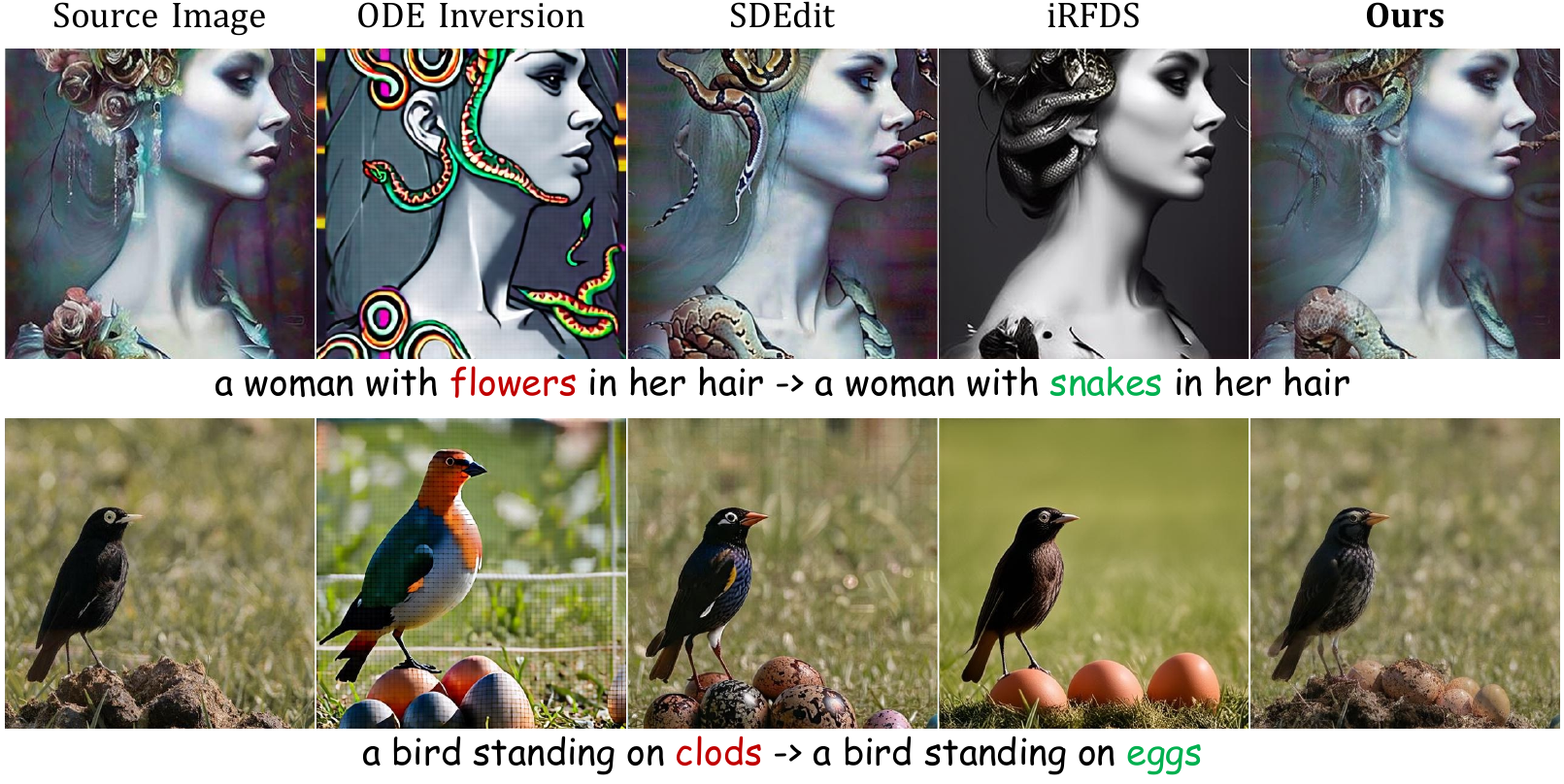}
        \caption{Text-based image editing comparisons with three baselines on PIE-Bench}
    \label{fig:10}
\end{figure*}

We ablated $\alpha$ in the initialization weight of $\lambda=\sigma^\alpha$ and provide visualizations in Figure~\ref{fig:8}. Besides, we provide additional qualitative comparisons with SDEdit~\citep{meng2021sdedit} (\cf, Section~\ref{sec:5.1}) on four image restoration tasks in Figure~\ref{fig:6} and Figure~\ref{fig:13}. Then, we provide additional qualitative comparisons with TTM~\citep{singer2025time} and GWTF~\citep{burgert2025go} (\cf, Section~\ref{sec:5.2}) on camera-controlled video generation tasks in Figure~\ref{fig:11}. Moreover, we gave more camera-controlled video generation results with Wan2.2 as the base model in Figure~\ref{fig:12}. Finally, we provided some text-based image editing cases of our method in Figure~\ref{fig:10}.

\newpage
\clearpage

\newpage
\clearpage

\newpage
\clearpage

\newpage
\clearpage

\newpage
\clearpage

\newpage

\end{document}